\documentclass{article}

\usepackage[utf8]{inputenc} 
\usepackage{url}            
\usepackage{booktabs}       
\usepackage{amsfonts}       
\usepackage{amsmath}
\usepackage{graphicx}
\usepackage[T1]{fontenc}    
\usepackage{hyperref}       
\usepackage{nicefrac}       
\usepackage{microtype}      
\usepackage{xcolor}         

\usepackage{textcomp}     

\newcommand{\etal}{\textit{et al.}~} %

    \PassOptionsToPackage{numbers, sort&compress}{natbib}

\usepackage[preprint]{neurips_2021}

\title{SyReNets: Symbolic Residual Neural Networks}
\author{%
  Carlos Magno C. O. Valle\\
  Technical University of Munich, Germany\\
  \texttt{carlos.valle@tum.de}\\
  \and
  Sami Haddadin\\
  Technical University of Munich, Germany\\
  \texttt{haddadin@tum.de} \\
}
  
\begin{document}
%

\maketitle

\vspace{-3pt}
\begin{abstract}
  Many research efforts have been put into modeling physical systems dynamics using data-driven methods. Despite successful seminal works on passive systems in the literature, learning free-form physical laws for controlled dynamical systems given experimental data is still an open problem. With precise models, complex behaviors can be predicted, generalized, planned, and explored with significant accuracy. For decades, symbolic mathematical equations and system identification were the golden standards. Unfortunately, a set of assumptions about the properties of the underlying system is required, which makes the model very rigid and unable to adapt to unforeseen changes in the physical system, i.e., there is no individual strategy that can be applied to hard-to-model or shape-shifting systems without additional incorporation of specific prior knowledge about it. Neural networks, on the other hand, are known universal function approximators but are prone to over-fit, limited accuracy, and bias problems, which makes them alone unreliable candidates for such tasks. In this paper, we propose \textit{SyReNets}, an approach that leverages neural networks for learning symbolic relations to accurately describe dynamic physical systems from data. It explores a sequence of symbolic layers that build, in a residual manner, mathematical relations that describes a given desired output from input variables. We apply it to learn the symbolic equation that describes the Lagrangian of a given physical system. We do this by only observing random samples of position, velocity, and acceleration as input and torque as output. Therefore, using the Lagrangian as a latent representation from which we derive torque using the Euler-Lagrange equations. The approach is evaluated using a simulated controlled double pendulum and compared with a vanilla implementation of neural networks, genetic programming, and traditional system identification. The results demonstrate that, compared to neural networks and genetic programming, \textit{SyReNets} converges to representations that are more accurate and precise throughout the state space. Despite having slower convergence than traditional system identification, similar to neural networks, the approach remains flexible enough to adapt to an unforeseen change in the physical system structure.  
  
\end{abstract}

\vspace{-3pt}
\section{Introduction}
\vspace{-3pt}
\label{section:in}

Neural networks have successfully been applied in a complex range of hard to solve problems, e.g. convolutional neural networks \cite{krizhevsky2012imagenet}, generative adversarial networks \cite{goodfellow2014generative} and transformers \cite{vaswani2017attention}, respectively changed their sub-fields of research. Yet, there exist areas that are still barely influenced by neural architectures. In this paper, we are particularly interested in modeling dynamics of physical systems. The importance of having a reliable dynamic model is clear for feed-forward motion generation, for motion planning, and when a swift dynamic response is required. It also allows for more accurate simulations of the underlying system, which in turn permits longer horizon of motion predictions. Additionally, the controller would be able to work with smaller gains to achieve a given state, which leads to less stiff actuators that are safer to interact with \cite{good1985dynamic, an1989role}. 

For many decades, the standard approach for modeling the dynamics of a given physical system is done by measuring or estimating the actuation positions, velocities, accelerations and finding relations between the commanded inputs (torque for example). Those relations would then be formulated by a set of mathematical equations that describes the motion of the system \cite{lynch2017modern}. 
As an example, Equation~\ref{eq_inverse_dyn} represents the inverse dynamics of a serial-chain rigid-body robotic system:
\begin{equation}\label{eq_inverse_dyn}
    \mathbf{\tau} = M(\mathbf{q})\ddot{\mathbf{q}} + C(\mathbf{q},\dot{\mathbf{q}})\dot{\mathbf{q}} + G(\mathbf{q})
\end{equation}
It describes the relation between torque ($\mathbf{\tau}$) and position, velocity, and acceleration ($\mathbf{q}, \dot{\mathbf{q}},\ddot{\mathbf{q}}$) at joint-level. $M$, $C$ and $G$ are pre-formulated matrix functions that mathematically describe inertial forces, centrifugal and Coriolis forces, and gravitational forces, respectively, in terms of mass, center of mass, and inertial matrix of each rigid-body. Traditional system identification methods aim to approximate them and, therefore, is still used due to its mathematical stability in the whole observable state space of the agent. The disadvantage is that the equations have fixed terms and even if some of the numerical parameters can be changed over time it limits the physical system to a given set of assumptions (e.g. shape and actuation type), making it complex to optimize for systems made out of hard to model materials (e.g. rubber) or that might be subject to unforeseen changes in structure for different applications.

Neural networks are generally known to be universal function approximators \cite{hornik1989multilayer} but they are prone to over-fit to the training data and have biased decisions when trained with an imbalanced or incomplete dataset \cite{wang2016training, geman1992neural, buda2018systematic}. Therefore, they, alone, are not reliable enough to be a surrogate model. However, for most dynamical systems belonging to our specific use case, there exists an analytical exact mathematical solution that can accurately describe the dynamics. For those, a different class of algorithms can be applied. For a lack of better term, we denote them as universal exact function estimators since instead of approximating a function they seek to estimate the exact underlying one that solves the problem. The most well known representative of this class is genetic programming \cite{koza1992genetic}, which is an evolutionary strategy for evolving symbolic tree representations. Each tree encodes a symbolic function. There, several generations of candidate symbolic solutions are tested and combined until a convergence criterion is reached. Despite suffering the same problems as neural networks, since those methods are mostly not dependent on gradient, if an underlying function exists it can eventually be found due to the random nature of the evolutionary process, given enough time and re-initializations. However, the search is highly inefficient since the process is fundamentally dependent on randomness, leading to repetitive and "unnecessary" evaluations of candidate solutions. Therefore, those approaches tend to not scale well with the increase of dimensionality of the problem \cite{bellman1966dynamic}. The truth is that exact symbolic estimation is inherently an NP-hard problem, which explains why dynamic model learning is still an open problem to solve.

We argue that a hybrid approach is a reasonable way to obtain the best of both worlds, namely the expressiveness of neural network architectures and the stability potential of symbolic representations. This would allow a general symbolic learning architecture to be applicable to any physical system that can be expressed analytically, potentially mitigating the disadvantages of "approximation" algorithms and the search inefficiency of the exact estimation methods. 

In this paper, we present Symbolic Residual Neural Networks (\textit{SyReNets}), a neural network architecture capable of learning symbolic mathematical relations from streams of data. \textit{SyReNets} can potentially be applied to estimate a vast number of functions, describing dynamics of many different physical systems. Traditionally, most methods try to learn the inverse dynamics (as denoted by Equation~\ref{eq_inverse_dyn}), since it conveys the relation between position, velocity and acceleration to applied torques \cite{rueckert2017learning}. However, each actuator adds one function to the number of equations to learn. Aiming at learning a fix and minimal number of equations we chose to learn the Lagrangian representation of the system:
\begin{equation}\label{eq_lagrangian}
    L = T - V
\end{equation}
Where, $L$ represents the Lagrangian, $T$ is the kinetic energy and $V$ is the potential energy of the system. By construction, it is always in $\mathbb{R}$, making this a hyper-compressive representation that maps from $\mathbb{R}^n \rightarrow \mathbb{R}$ for $n$ inputs.
From the Lagrangian equation, it is possible to derive the inverse dynamics of the system using the Euler-Lagrange method:
\begin{equation}\label{eq_euler_lagrangian}
    \mathbf{\tau}_i = \frac{\mathrm{d}}{\mathrm{d}t}\frac{\partial L}{\partial \dot{\mathbf{q}}_i} - \frac{\partial L}{\partial \mathbf{q}_i}
\end{equation}
This operations transform the Lagrangian to $m$ possible actuator torques, effectively constructing a map from $\mathbb{R} \rightarrow \mathbb{R}^m$. Therefore, the Lagrangian can be considered a highly compact latent representation to a set of non-injective functions.

The paper is structured as follows: In Section~\ref{section:rw} we will discuss related works that take advantage of energy representations to approximate or estimate physical systems, highlighting what they might be lacking to be of general use. In Section~\ref{section:me}, we present our method in detail, followed by Section~\ref{section:ex} where we evaluate the performance of the approach on a simulated double pendulum system against vanilla neural networks, genetic programming and traditional system identification. 
Finally, in Section~\ref{section:co} we present our conclusions from the experiments, possible future works and potential societal impacts of our method.

\vspace{-3pt}
\section{Related Works}
\vspace{-3pt}
\label{section:rw}
Related works on learning dynamic representations for physical systems from data can be categorized into two approaches (as briefly described in Section~\ref{section:in}):
\subsection{Approximation Approaches}

In \cite{cranmer2020lagrangian}, authors propose a neural network architecture for learning a parametric Lagrangian which they use to compute the Euler-Lagrange equations by using an automatic differentiation procedure. They show that it is possible to use the black-box representation of the Lagrangian to obtain estimates for the acceleration of the system and use those to predict a given trained system behavior just by having an initial position and velocity. They start from the assumption that the physical system is not actuated since the energy is stationary (the left-hand side of (\ref{eq_euler_lagrangian})). This limits the applicability of the method. This problem is partially solved in \cite{lutter2019deep}, where the authors propose to learn the Lagrangian of the system with variable torque, they use it to build the dynamic equations. Unfortunately, they assume the system is composed of rigid structures, limiting once more the applicability of the approach.

The work of \cite{NEURIPS2019_26cd8eca} proposes a similar approach to the previous one but using the Hamiltonian instead. They take advantage of the Hamiltonian equations to describe the relation between position, momentum, and Hamiltonian. The problem with their approach is that it depends on having momentum data, which is not always available and, leads to additional problems as described in \cite{cranmer2020lagrangian}.


\vspace{-3pt}
\subsection{Exact Estimation Approaches}
\vspace{-3pt}

In the seminal work of Schmidt and Lipson \cite{Schmidt81}, genetic programming \cite{koza1992genetic} was used to evolve a set of equations for exact symbolic estimation of unknown physical systems.
In each generation, they compute the partial derivative of pairs of variables and compare to numerical partial derivatives, if the equality is within a given tolerance error the equation is saved, in that way a dictionary of valid equalities is built. By observing the Pareto frontier on the error and function complexity, several correct mathematical relations can be obtained. The problem with this approach is that it does not scale well to more complex problems due to the higher complexity of the search space.

To mitigate this issue, Udrescu \etal \cite{NEURIPS2020_33a854e2}
recently proposed a state of the art symbolic regression method that search for pareto-optimal equations from data using a combination of brute-force search, neural networks approximation and graph modularity simplification in order to alleviate the curse of dimensionality problem. They argue that their method is orders of magnitude more robust to noise and bad data in comparison to the previous state of the art. But they experiment only with datasets with directly measured outputs, which is not the goal with our work since we cant obtain measures for the Lagrangian. 
\begin{figure}[!htb]
  \centering
  \fbox{\rule[-.5cm]{0cm}{0cm} 
  \includegraphics[width=0.9\linewidth]{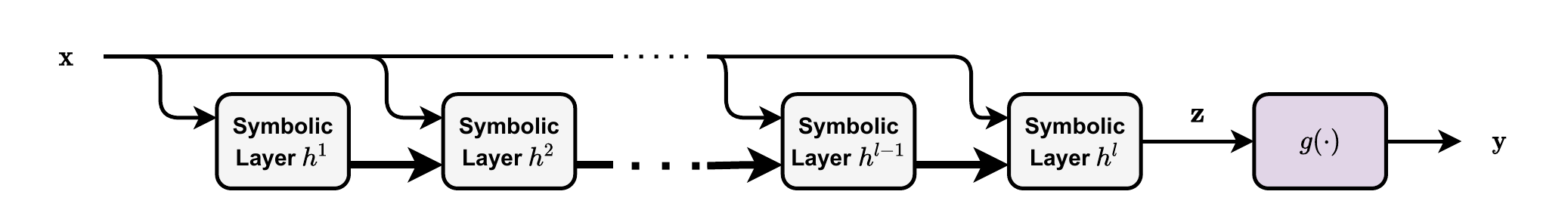}
  \rule[-.5cm]{0cm}{0cm}}
  \caption{\textit{SyReNets} high-level architecture}
  \label{fig_high_level}
\end{figure}
\vspace{-3pt}
\section{Methodology}\label{section:me}
\vspace{-3pt}
We start from the minimum set of assumptions that the system of interest is accurately describable by a mathematical equation of the form $f(\mathbf{x}) = \mathbf{z}$, where $f(\cdot)$ denotes the equation to be learned which maps an input $\mathbf{x}$ into an output $\mathbf{z}$. We also assume that $\mathbf{z}$ is not necessary directly measurable (especially in our use case, where we intend to learn the Lagrangian), meaning that there is a known function $g(\mathbf{z}) = \mathbf{y}$, that maps $\mathbf{z}$ to $\mathbf{y}$, which is our observable variable. Therefore:
\begin{equation}\label{eq_x1.2}
\begin{aligned}
f(\mathbf{x}) = \mathbf{z} ;    g(\mathbf{z}) = \mathbf{y}
\end{aligned}
\end{equation}
with $g(\cdot)$ being a known injective or non-injective transformation function. For our use case $g(\cdot)$ is the Euler-Lagrange transformation, which is non-injective. 

We assume $f(\cdot)$ can be defined as an aggregation or composition of less complex components. Those, can depend on the original function input $\mathbf{x}$ or other sub-components. Therefore, there exists a natural hierarchical structure that can generate complex mathematical relations from simpler operations. This relation is evident in:
\begin{equation}\label{eq_complex_aggregation}
f(\mathbf{x}) = h^l(\mathbf{x}, h^{l-1}(\mathbf{x}, h^{l-2}( ... h^{1}(\mathbf{x})))) 
\end{equation}
Where $h^i(\mathbf{x}, h^{i-1})$ is the $i^{th}$ function that applies a given set of operations to the input $\mathbf{x}$ and the previous function $h^{i-1}$. 

Our proposed solution, \textit{SyReNets}, takes advantage of this modular property assumption and, therefore, is made of a sequence of simple symbolic layers, which are composed to obtain a more complex equation. Figure~\ref{fig_high_level} depicts the architecture on a high level. Each symbolic layer receives a set of samples of the original input $\mathbf{x}$ and the output of the previous layer. This allows a residual inspired behavior \cite{He_2016_CVPR} where new symbolic relations with $\mathbf{x}$ can be generated at every layer and not exclusively dependent on the outputs of the previous one. The input samples are randomly drawn from a uniform distribution $\mathcal{U}(-\tfrac{\pi}{2}, \tfrac{\pi}{2})$ in order to cover the majority of the state-space. Since we are learning the Lagrangian of a physical system there is usually no inherit time-dependence, therefore, there is no need to sample data sequentially.

In Figure~\ref{fig_representation_detailed} its possible to see the high-level structure of each symbolic layer. They are composed of two main parts, a representation transformation block, and $k$ selection heads. 

\vspace{-3pt}
\subsection{Representation transformation}
\vspace{-3pt}

\begin{figure}
  \centering
  \fbox{\rule[-.5cm]{0cm}{0cm} 
  \includegraphics[width=0.9\linewidth]{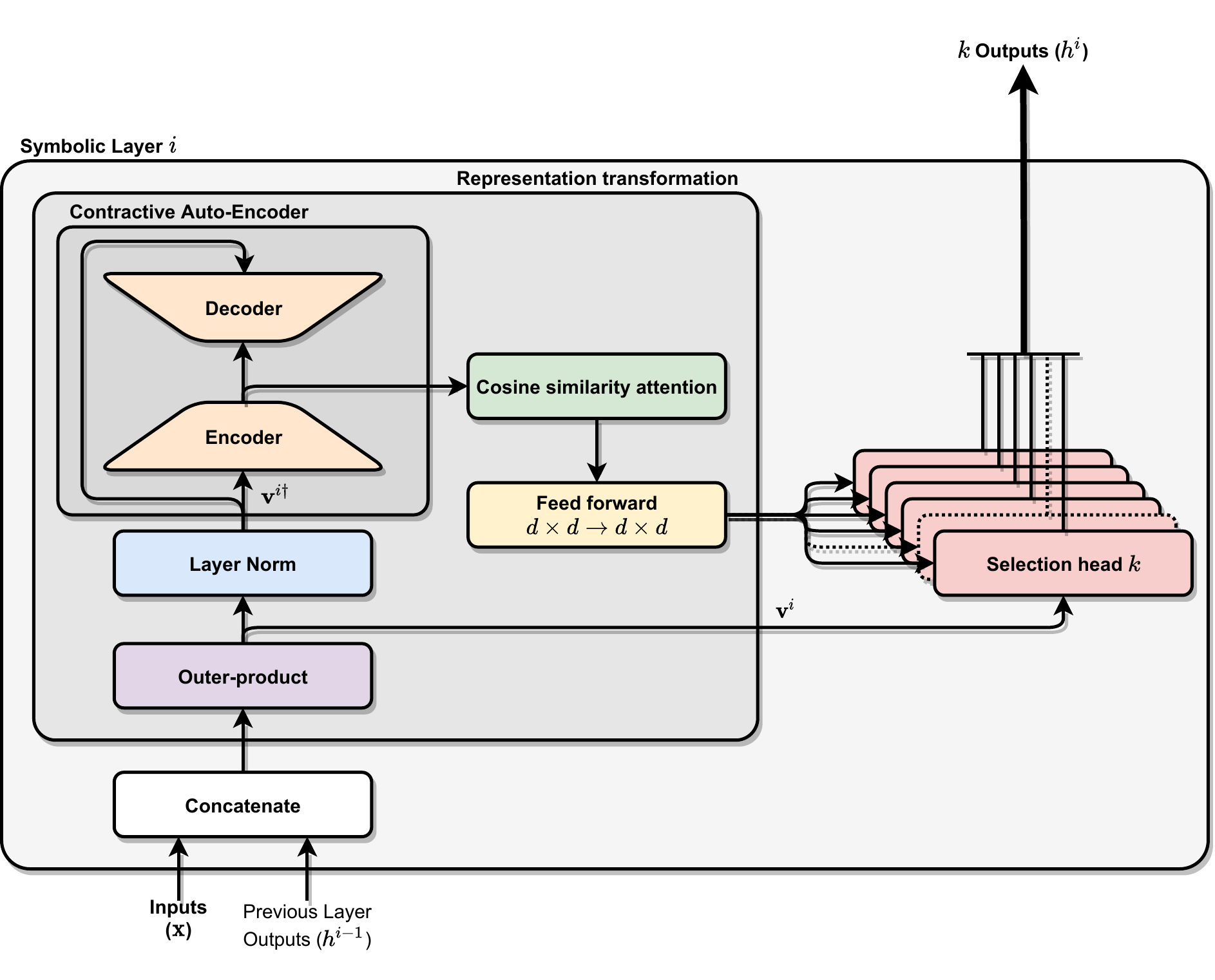}
  \rule[-.5cm]{0cm}{0cm}}
  \caption{Detailed representation block of the symbolic layer }
  \label{fig_representation_detailed}
\end{figure}

The transformation block changes the inputs given to the layer to a representation that conveys the relation between the input and possible mathematical transformation that can be applied to it. Figure~\ref{fig_representation_detailed} illustrates the process. After concatenating the input $\mathbf{x}$ with the previous layer output $h^{i-1}$, the first sub-component is responsible for generating all possible unique combinations of the input samples using a given set of mathematical operators. In the scope of this work, we explored the following operators: $(+,\times,\sin, \cos)$. Those were specially chosen because they are sufficient to represent our use case but many other mathematical operations could be used here. The arity and symmetry of the operation are essential for an efficient calculation, for unary operations we only need to apply it to all input samples, but for binary we compute the outer product of the input samples using said operator, if it is symmetric we care mostly for the upper (or lower) triangle, if not the full matrix will be required. We did not explore higher arities in this work but an equivalent relation probably exists. We then collect all unique terms in a vector $\mathbf{v}$ which represents all candidate changes that this specific layer can do to the input.  

In the next sub-block, since we do not necessarily know if the state-space of our output is the same one as the observed output, We apply a contractive auto-encoder \cite{rifai2011contractive}, composed by the encoder $f_{ae}(\cdot)$ and the decoder $g_{ae}(\cdot)$. It embeds samples of the full vector of candidate changes into a latent representation of size $d$ that should be more compact, robust by design, and aid in the selection process, this last relation will be clearer in Section~\ref{subsec_loss}. Normalization of inputs improves performance of the auto-encoder, therefore, the inputs are normalized and denoted by $\mathbf{v}^\dagger$. We apply layer normalization \cite{ba2016layer} because the only information we can assume is that all auto-encoder inputs at a given sample are from the same point in the state-space of $\mathbf{x}$, therefore, we can find relations in that dimension but no assumption can be made regarding the order of samples. The same auto-encoder is shared between all symbolic layers to ensure that the latent representation is not dependent on the specific layer it is used. 

The encoded outputs are then used to compute the cosine similarity between all samples, this serves to observe similarities between candidate equations without assuming any particular order of samples, in a similar way as observed in \cite{vaswani2017attention}. The symmetric matrix generated of size $d\times d$ at this point is independent of the symbolic layer, since every block until now is shared between all layers. To allow the architecture to specify for each symbolic layer we added a single fully connected neural network layer with linear activation that transforms the symmetric matrix row by row, therefore, observing the similarity between all candidates to a particular one in each operation. This operation denotes a local representation with the same dimensions as before but conveying understanding of how a particular candidate relates to the others, this information is then fed to all selection heads.

\vspace{-3pt}
\subsection{Selection heads}
\vspace{-3pt}

\begin{figure}
  \centering
  \fbox{\rule[-0cm]{-0cm}{0cm} 
  \includegraphics[width=0.9\linewidth]{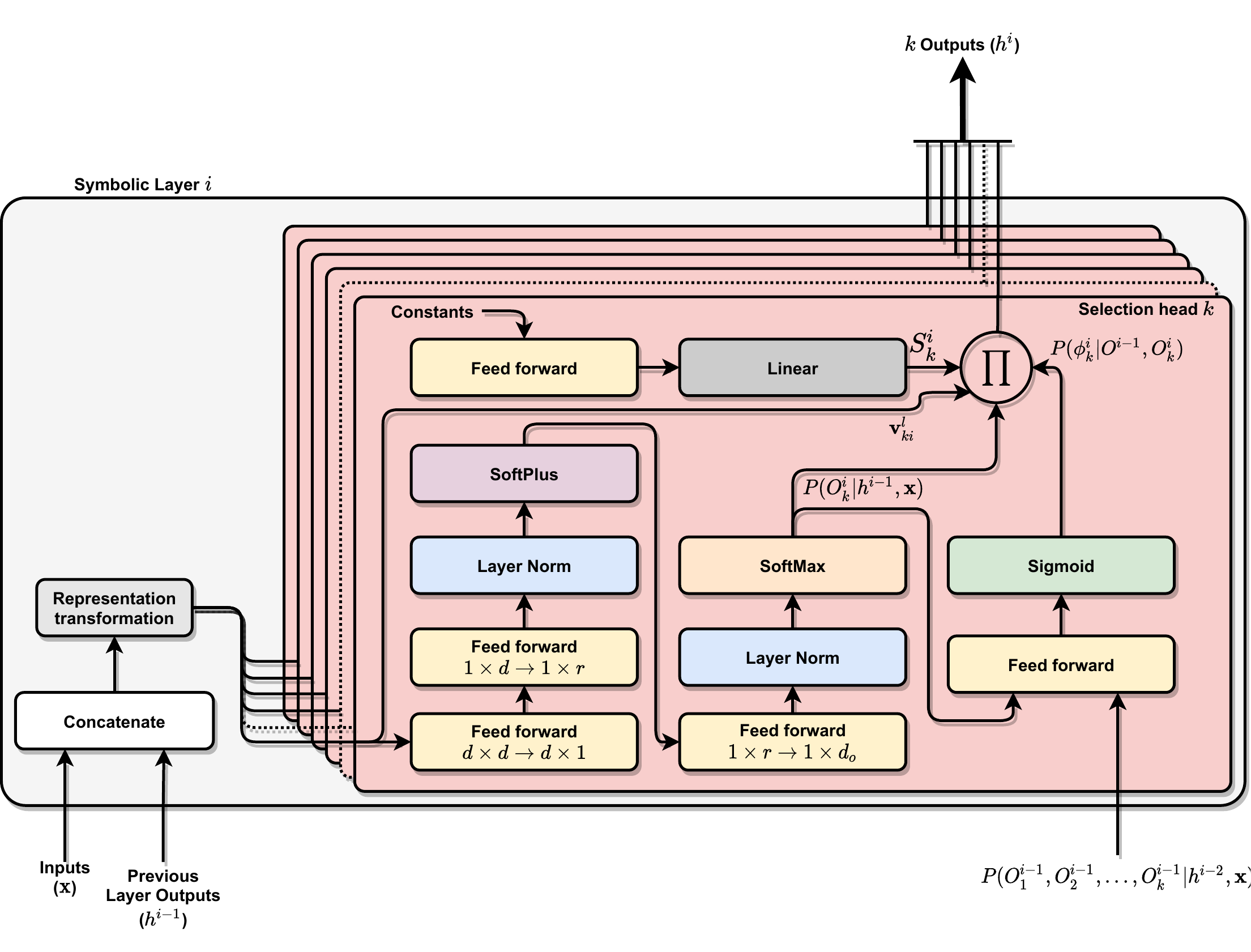}
  \rule[-0cm]{-0cm}{-10cm}}
  \caption{Detailed selection head block of the symbolic layer}
  \label{fig_selection_head_detailed}
\end{figure}

Each selection head is responsible for selecting one operation $h_j^i$ to flow to the next layer, where $j$ is the $j^{th}$ selection head if the $i^{th}$ symbolic layer. Figure~\ref{fig_selection_head_detailed} illustrates the sub-modules of this block, the first operation computes a weighted sum of the input matrix, , converting $d\times d\rightarrow d\times 1$ this is achieved with a single output single layer neural network with no bias and linear activation. This operation allows each head to focus on specific sub-groups of candidate solutions. 

The next sub-module is responsible for the actual selection of the candidate, this is done with a fully connected neural network with one hidden layer with $r$ neurons with softplus activation and an output layer with $d_o$ neurons with softmax activation, all layers without bias but with layer normalization before the activation function. The components of this sub-module were selected targeting minimizing the number of parameters in each head without potentially damaging the performance of the selection head. So this block may be subject to variation depending on the complexity of the problem. The output of this block is a probability distribution $P(O_i|h^{i-1},\mathbf{x})$ of selecting a given candidate solution. If we chose to sample a candidate from this distribution we would lose all the information regarding the gradients of the other candidate solutions since they would become zero. So we opted to use the full probability distribution as part of the output of each selection head.

To complement, since probability distributions are naturally bounded between 0 and 1, we need to add a scalar operation $S$ that would allow the selected operation to virtually have no bound. We used a single-layer fully connected neural network with constant inputs $c$ and $d_o$ outputs, for this purpose. As a final modification we imbued the selection head with the ability to control if its output should be used or not. We denominate this mechanic as on-off probability gate $\phi$, similar to the forget mechanism in long short-term memory units \cite{HochSchm97} and is described by a single-layer fully connected neural network with sigmoid activation and no bias. It receives as inputs the current values for the head selection probability distribution ($P(O_j^{i}|H^{i-1},\mathbf{x})$) and the joint probability of the last layer selection probability distribution ($P(O_1^{i-1},O_2^{i-1},...,O_k^{i-1}|h^{i-2},\mathbf{x})$). If we ensure that every head has independent probabilities to each other we can easily apply to product rule, therefore:
\begin{gather}
    P(O_1^{i-1},O_2^{i-1},...,O_k^{i-1}|h^{i-2},\mathbf{x}) = \prod_j{P(O_{j}^{i-1}|h^{i-2},\mathbf{x}}) \\
    P(\phi_{j}^i|O^{i-1}, O_j^i) = \sigma(W_1 P(O_{k}^{i}|h^{i-1},\mathbf{x})+W_2 \prod_j{P(O_{j}^{i-1}|h^{i-2},\mathbf{x}}))
\end{gather}
where $\sigma$ is the sigmoid activation function and $W_1$ and $W_2$ are the weights of the neural network. Finally, all the previously mentioned components are aggregates in the following equation:
\begin{equation}\label{eq_selection_head_out}
h_j^i = P(\phi_{j}^i|O^{i-1}, O_j^i) \sum_b{S_{jb}^i \mathbf{v}_{b}^i P(O_{jb}^i|h^{i-1},\mathbf{x})}
\end{equation}
Equation~\ref{eq_selection_head_out} represents the output of the $j^{th}$ selection head of the $i^{th}$ layer. The sum over $b$ represents the sum over all the discrete probability bins representing candidate operations, this is possible because we have a finite number of possible combinations to select from.

All subsequent layers have the same structure with exception to the $l^{th}$ layer, which is the last one. This one sums all selection heads outputs into a single output, as described by: 
\begin{equation}\label{eq_sum}
    \hat{f}(\mathbf{x}) = \sum_j{h_j^l}
\end{equation}
\vspace{-3pt}
\subsection{Loss function}
\vspace{-3pt}
\label{subsec_loss}

The output of the SyReNet architecture is a scalar value generated by the learned estimated symbolic equation $\hat{f}(\mathbf{x})$. In our application it represents the Lagrangian of the learned system. Since we can not easily measure energy, we need to convert it to some measurable magnitude. We, therefore, use the Euler-Lagrange method as a transformation function $g(\cdot)$ to map the equivalent values in torque. The basic loss function $\mathcal{L}_{b}$ is the mean squared error (MSE) between the desired torque and the value obtained by the model, however, this loss alone is not enough for our particular architecture. We also need to add the loss terms $\mathcal{L}_{ae}$ obtained from the auto-encoder.
\begin{equation}\label{eq_loss_basic}
\mathcal{L}_{b} = \frac{1}{N}\sum_i^N{(y_i - g(\hat{f}(\mathbf{x}_i)))^2} + \mathcal{L}_{ae}    
\end{equation}
\vspace{-5pt}
where,
\begin{equation}\label{eq_loss_ae}
\mathcal{L}_{ae} = 
\frac{1}{N}\sum_i^N{[(\mathbf{v}^{i\dagger} - g_{ae}(f_{ae}(\mathbf{v}^{i\dagger})))^2 + \lambda_1 \left \| J_{f_{ae}}(\mathbf{v}^{i\dagger}) \right \|_F^2]}    
\end{equation}
where $\lambda$ is a Lagrangian multiplayer that weights the importance of of this extra term in the loss and $J_{f_{ae}}(\cdot)$ is the Jacobian of the non-linear mapping  of the encoder, as stated in \cite{rifai2011contractive}.
The auto-encoder is an essential part of the architecture. It allows the attention sub-component to search for similarity in a latent representation that is in a state space that has a higher probability to be related to $\mathbf{y}$ in (\ref{eq_x1.2}) since the gradients flowing back from the $\mathcal{L}_{b}$ adjust all learnable weights in order to minimize. If the auto-encoder was not used we would search for similarity directly in the input variable state space which might have a non-linear relation to $\mathbf{y}$. 

However, some challenges require additional loss terms. One of them is that each symbolic layer is composed of multiple independent selection heads, since there is no direct communication between them it is possible to assume that, in some cases, they will converge to the same candidate solution. For each layer, we want each head, $O_i$, distribution to be independent from one another, a clear way to achieve this is to add a cross-entropy loss $H(\cdot ,\cdot)$ between each different head to be maximized. Additionally, entropy $H(\cdot )$ should be minimized, allowing simpler formulas to be discovered since lower entropy increases the likelihood of a single equation selected by each head. A complementary issue is that every time one of the heads changes too much its probability distribution it introduces disturbances in the whole cascade of layer after it. To mitigate this, we multiplied the $\phi_{i}$ gate with the $O_i$ distribution before computing the previously mentioned operations. The idea behind it is to give the network more flexibility to shift the distribution when $\phi_{i}$ is low, generating something similar to a null-space. Equation~\ref{eq_loss_complementary} formulates the above stated complementary loss $\mathcal{L}_{c}$.
\begin{equation}\label{eq_loss_complementary}
\mathcal{L}_{c} = \sum_{i=0}^{l}\sum_{j=0}^{k}\lambda_2 H(p_j^{i}) -\lambda_3 \sum_{{j}'\neq j}^{}{H(p_{{j}'}^{i},p_j^{i})}
\end{equation}
where,
\begin{equation}
p_{j}^i = P(\phi_{j}^i|O^{i-1}, O_j^i)P(O_{j}^i|h^{i-1},\mathbf{x})) \\
\end{equation}
the final loss is denoted by
\vspace{-1pt}
\begin{equation}\label{eq_loss_total}
\mathcal{L} = \mathcal{L}_{b} + \mathcal{L}_{c}   
\end{equation}




\vspace{-3pt}
\section{Experiments}
\vspace{-3pt}
\label{section:ex}
We evaluated the approach in a simulated double pendulum system, which can be verified analytically. We aim to evaluate the convergence precision and accuracy of the method by comparing it to representative approaches from each one of the groups presented in section~\ref{section:rw} from the literature, namely, neural network for approximation methods and genetic programming for exact estimation. System identification was also evaluated as a control measure to compare how far the performance is to the current most reliable method. 

The system under evaluation is dependent on 2 actuated joints, where the $i^{th}$ joint torque $\tau_i$ (outputs), inverse dynamics dependents on $q$, $\dot{q}$ and $\ddot{q}$ (inputs). The equation that describes the Lagrangian $L$ of this system is:
\begin{multline}\label{ex_lagrangian_double_pend}
L = \tfrac{1}{2}(\tfrac{m_1}{3} + m_2)l_1^2\dot{q}_1^2 + \tfrac{1}{2}(\tfrac{m_2}{3})l_2^2\dot{q}_2^2 + \tfrac{1}{2}m_2l_1l_2\dot{q}_1\dot{q}_2\cos{(q_1-q_2)} \\
+(\tfrac{m_1}{2} + m_2)gl_1cos(q_1) +(\tfrac{m_2}{2})gl_2cos(q_2)    
\end{multline}
where $g$ is the gravity acceleration, $m_i$, $l_i$ are the mass and length of the $i^{th}$ link and $q_i$, $\dot{q}_i$ and $\ddot{q}_i$ are position, velocity and acceleration for the $i^{th}$ joint. For our experiments we defined the values of $m_1 = 3.0$, $l_1 = 2.67$, $m_2=1.0$, and $l_i=1.67$, there is no particular reason for selecting those values. 

As stated in Section~\ref{section:me}, we use a uniform distributed $\mathcal{U}(-\tfrac{\pi}{2}, \tfrac{\pi}{2})$ input sampling from all joints $q$, $\dot{q}$ and $\ddot{q}$. We calculate numerically their respective $L$ values using (\ref{ex_lagrangian_double_pend}) and we compute the torque at each joint $\tau_i$ with (\ref{eq_euler_lagrangian}), which represents our transformation function $g(\cdot)$. We used the automatic differentiation package from the Pytorch Python library \cite{NEURIPS2019_9015} to numerically calculate the required derivatives of each model.

We sampled $32000$ sets of data points for training, subdivided into mini-batches of 32 samples. For the test data-set we sampled 10000 sets of points. For the genetic programming tests we used the GPTIPS 2.0 \cite{searson2015gptips} toolbox (subject to the GNU public software license, GPL v3) in Matlab 2017b. The neural networks, system identification and the \textit{SyReNets} implementation were developed in Python 3.6.13 using the Pytorch library version 1.8.1 (under BSD license), the \textit{SyReNets} implementation will be made publicly available in the near future. 

For our evaluation hardware, the GPTIPS 2.0 toolbox used an Intel Core i9-9900KF CPU while systems implemented using Pytorch were optimized to run on GPU and used an RTX 2080Ti. Therefore, it is not fair to compare execution times with the genetic programming approach. For each experiment a set of 10 random seeds were used, we will evaluate each method's performance in three groups: the 5 best performing ones, representing the best you can expect; all 10 seeds, presenting the average performance; and the 5 worst performing ones.  

The genetic programming representative is based the multi-gene genetic programming algorithm (MGGP) as described in \cite{gandomi2012new} with population size of 1000 candidates, ran for 300 generations, using tournament of size 15 as selection and elite fraction of 10\%. The operations provided to drive the evolutionary process where the binary operations of $+$, $-$, $\times$ and the unary operations $\sqrt{(\cdot)}$, $(\cdot)^2$, $(\cdot)^3$, $\sin{(\cdot)}$, and $\cos{(\cdot)}$.
The execution was terminated if the maximum number of iterations were reached or if the MSE of the best performing solution was lower than $10^{-10}$.

The neural networks (NN) were implemented with five hidden layers with $300$ neurons each with softplus activation function and one output layer with one neuron with linear activation. The loss was also the MSE. The network was trained via stochastic gradient descent using Adam optimizer \cite{kingma2014adam} with $\beta_1=0.9$, $\beta_2=0.999$, and initial learning rate of $10^{-3}$ which was divided by 10 for every 2000 iterations without any improvement in the loss up to $10^{-5}$. The step with the lowest training loss is saved until a better performance is achieved or a time limit is reached, when the training is stopped.

The \textit{SyReNets} were implemented with 3 symbolic layers, each with 12 selection heads. The auto-encoder had 2 hidden layers with 128 neurons and softplus activation, both in the encoder and decoder. The output of the encoder had 16 neurons with linear activation. The hidden state of the selection head had 64 neurons, any other input to the selection head had the same number of inputs which was the number of candidate solutions $\mathbf{v}$. With exception to the auto-encoder, none of the neural network layers used bias. The constant inputs for the scaling function were a vector of ones. the loss terms $\lambda_1=\lambda_3=1$ and $\lambda_2=0.001$. The architecture was also trained with Adam with the same hyper-parameters as stated previously but the learning rate decay was performed at every 1000 non-improving steps. We also store the best performing state of the system until a time limit is reached and the training stops or a better performance is obtained.

The system identification (SysId) setup had a symbolic form of the equation where it had to approximate the values for $m_1$, $l_1$, $m_2$, and $l_2$ using a neural network with a constant input vector of ones. The network had one hidden layer with 64 neurons using softplus activation with no bias and one output layer with 4 outputs using linear activation. The training used Adam with the same parameters as before but with no decaying learning rate, which was kept constant at $10^{-3}$. During training, similar to the previous ones, the parameters of the best performing state on the training data are stored until a training time limit is reached 

\vspace{-3pt}
\subsection{Learning from direct measurements}
\vspace{-3pt}
We first experimented on the capability of the method to learn directly from $L$ values, i.e., without using any $g(\cdot)$. In Table~\ref{tb_direct}, we can observe the performance of the method. All methods were trained for 2000 seconds except for the MGGP, which trained until the last generation for all seeds. It is clear that the best performing methods overall are SysId and MGGP that managed to achieve the exact equation. However, SysId  benefits from that it has information regarding the structure of the equation to learn, nevertheless, when we observe the worse performances we can see that some times both SysId and MGGP are not converging to the correct values, denoting that there exist some well-known sensitivity on those methods. Since SysId is fairly quick to converge (around 2 minutes in our experiments) this problem can be easily circumvented by repeating the experiments. Also, it might also be possible that the model used inside the SysId is inaccurately representing the system to be learned, therefore, it might not be able to converge at all. Those problems were not observed with NN and \textit{SyReNets}, which presented consistent performance with all sets of seeds. \textit{SyReNets}, in special, presented up to $5000\times$ better accuracy when compared to traditional neural networks which denotes the potential of symbolic approaches as surrogate models for physical systems. Unfortunately, the approach was not precise enough to estimate the exact equation as MGGP did, this is partially due to the probabilistic nature of the architecture's selection head which prevents that candidate operations selection probabilities becomes zero and, therefore, stopping the gradient flow through the function.     

\vspace{-3pt}
\subsection{Learning from indirect measurements}
\vspace{-3pt}
Our second experiment evaluates the capability of the architectures to learn indirectly the Lagrangian by observing the torque values as output of the transformation function $g(\cdot)$ which computes the Euler-Lagrange equations. In Table~\ref{tb_indirect} we can see that this time the MGGP algorithm could not converge to the correct equation, denoting that exact function estimators might have a limitation. The performance of neural networks and \textit{SyReNets} once more proved to be stable for all tested seeds, confirming the results from the previous experiment and proving their indirect learning capability. 

\begin{table}[]
\tiny
  \caption{MSE of Learning the Lagrangian directly for 2000 seconds}
\begin{tabular}{rlllll}
\label{tb_direct}
Seeds        &              & MGGP                  & NN                       & \textit{SyReNets}                & SysId                                          \\ \hline
Best         & train        & $\mathbf{0.000}$                 & $0.534$                    & $1\mathrm{e}{-4}$         & $\mathbf{9.775\mathrm{e}{-20}}$                           \\ 
             & test         & $\mathbf{0.000}$                 & $0.781$                    & $1.4\mathrm{e}{-4}$       & $\mathbf{1.1\mathrm{e}{-18}}$                             \\ \hline
5 Best       & train        & $11.293 (\pm 15.469)$   &  $0.8264 (\pm 0.202)$      &$0.025 (\pm 0.031)$        & $\mathbf{1.730\mathrm{e}{-17} (\pm 1.887\mathrm{e}{-17})}$ \\
             & test         & $11.827 (\pm 16.199)$   & $0.6257 (\pm 0.118)$       &$0.0316 (\pm 0.047)$       & $\mathbf{5.976\mathrm{e}{-17} (\pm 6.33\mathrm{e}{-17})}$  \\ \hline
All 10       & train        & $16.152 (\pm 15.009)$  & $ 1.507 (\pm 1.037)$       &$\mathbf{0.207  (\pm 0.371)}$       & $362.152 (\pm 764.072)$                          \\
             & test         & $16.964 (\pm 15.619)$  & $1.212 (\pm 1.032)$    &$\mathbf{0.137  (\pm 0.184)}$       & $363.408 (\pm 775.727)   $                       \\ \hline
5 Worse      & train        & $21.010 (\pm 14.441)$  & $ 2.189 (\pm 1.104)$       &$\mathbf{0.434 (\pm 0.493)}$        & $724.304 (\pm 992.813) $                         \\
             & test         & $22.101 (\pm 14.851)$  & $1.798 (\pm 1.235)$    &$\mathbf{0.270 (\pm 0.212)}$        & $726.817 (\pm 1011.82)  $                          \\ \hline
\end{tabular}
\end{table}

\begin{table}[]
\tiny
  \caption{MSE of Learning the Lagrangian indirectly (torque error) for 2000 seconds}
\begin{tabular}{rlllll}
\label{tb_indirect}
Seeds        &              & MGGP                  & NN                       & \textit{SyReNets}                & SysId                                          \\ \hline
Best         & train        & $5.264$                 & $0.141$                    & $1\mathrm{e}{-3}$         & $\mathbf{0.000}$                           \\ 
             & test         & $5.211$                 & $0.179$                    & $1\mathrm{e}{-3}$       & $\mathbf{3.831\mathrm{e}{-27}}$                             \\ \hline
5 Best       & train        & $92.484 (\pm 85.669)$   &  $0.1556 (\pm 0.019)$      &$0.004 (\pm 0.002)$        & $\mathbf{5.793\mathrm{e}{-32} (\pm 1.041\mathrm{e}{-31})}$ \\
             & test         & $92.347 (\pm 87.541)$   & $0.162 (\pm 0.057)$       &$0.004 (\pm 0.003)$       & $\mathbf{1.123\mathrm{e}{-27} (\pm 1.572\mathrm{e}{-27})}$  \\ \hline
All 10       & train        & $428.741 (\pm 399.637)$  & $ 0.207 (\pm 0.080)$       &$\mathbf{0.011  (\pm 0.012)}$       & $88.290 (\pm 186.189)$                          \\
             & test         & $444.862 (\pm 423.320)$  & $0.192 (\pm 0.061)$    &$\mathbf{0.012  (\pm 0.014)}$       & $178.963 (\pm 383.053)   $                       \\ \hline
5 Worse      & train        & $764.998 (\pm 263.317)$  & $ 0.259 (\pm 0.085)$       &$\mathbf{0.020 (\pm 0.014)}$        & $176.581 (\pm 241.890) $                         \\
             & test         & $797.376 (\pm 291.327)$  & $0.221 (\pm 0.056)$    &$\mathbf{0.131 (\pm 0.016)}$        & $357.927 (\pm 500.073)  $                          \\ \hline
\end{tabular}
\end{table}


\vspace{-3pt}
\section{Conclusion}
\vspace{-3pt}
\label{section:co}
In this paper we explored ideas on how to incorporate symbolic reasoning into traditional machine learning techniques, namely, we investigated the development of a hybrid architecture capable of mixing the approximation capabilities of neural networks and the stability potential of symbolic mathematical representations to learn dynamic models of physical systems. We also investigated the potential of this approach in learning the Lagrangian of the system, which is an extremely compact latent representations that typically can not be directly measured in real world applications. We experimented with the approach in a simulated double pendulum system and evaluated the accuracy in the task of learning the Lagrangian, first from direct measurements of the Lagrangian as a baseline test and later from indirect measurements, more specifically from torque measurements. The experiments showed that, despite not converging to the exact equation, the performance was consistent in all random seeds observed. We compared the performance with standard implementations of neural networks, genetic programming and system identifications as a control measure. Experiments show that system identification and genetic programming are able to find exact solutions but are also prone to convergence problems that prevents the solution to be found in all the tested seeds. Neural networks presented reasonable convergence but performed worse on average than \textit{SyReNets}, showing that such hybrid approaches has a lot of potential.

The downside of our approach is that we are, currently, dependent on all gradient information of all candidate functions. Any attempt we tried to deal with zero gradients caused instability in convergence and therefore were considered an architectural problem. Future efforts will be invested in allowing the existence of zero gradients since it is a key point to get exact solutions. Another issue experienced in training was with the number of steps to converge, while neural networks presented consistent error reduction \textit{SyReNets} had more variance with regard to the number of steps to converge. We hypothesize it is related to the required symbolic depth required to express certain elements of the equation, therefore, further investigation on compositionality has to be made in the near future.  

Potentially, this research can help to automate safety-critical model changes in physical systems such as robots in industrial and service environments, in particular in flexible automation setups or domestic environments. Also, the algorithms could be used as a fault detection and identification systems that compute symbolic explanation for better subsequent decision making. However, in the long run, the specific use cases require the careful consideration of use, in particular in human observation scenarios or in decision systems for military applications since explainable symbolic representations of data streams give potentially more introspection into the observed data.

\begin{ack}
We thank Dr. Fan Wu for his helpful comments. We gratefully acknowledge the general support by Microsoft Germany. 

\end{ack}



{
\small
\bibliographystyle{unsrtnat}
\bibliography{bibliography}

\begin{thebibliography}{26}
\providecommand{\natexlab}[1]{#1}
\providecommand{\url}[1]{\texttt{#1}}
\expandafter\ifx\csname urlstyle\endcsname\relax
  \providecommand{\doi}[1]{doi: #1}\else
  \providecommand{\doi}{doi: \begingroup \urlstyle{rm}\Url}\fi

\bibitem[Krizhevsky et~al.(2012)Krizhevsky, Sutskever, and
  Hinton]{krizhevsky2012imagenet}
Alex Krizhevsky, Ilya Sutskever, and Geoffrey~E Hinton.
\newblock Imagenet classification with deep convolutional neural networks.
\newblock \emph{Advances in neural information processing systems},
  25:\penalty0 1097--1105, 2012.

\bibitem[Goodfellow et~al.(2014)Goodfellow, Pouget-Abadie, Mirza, Xu,
  Warde-Farley, Ozair, Courville, and Bengio]{goodfellow2014generative}
Ian Goodfellow, Jean Pouget-Abadie, Mehdi Mirza, Bing Xu, David Warde-Farley,
  Sherjil Ozair, Aaron Courville, and Yoshua Bengio.
\newblock Generative adversarial nets.
\newblock In Z.~Ghahramani, M.~Welling, C.~Cortes, N.~Lawrence, and K.~Q.
  Weinberger, editors, \emph{Advances in Neural Information Processing
  Systems}, volume~27. Curran Associates, Inc., 2014.
\newblock URL
  \url{https://proceedings.neurips.cc/paper/2014/file/5ca3e9b122f61f8f06494c97b1afccf3-Paper.pdf}.

\bibitem[Vaswani et~al.(2017)Vaswani, Shazeer, Parmar, Uszkoreit, Jones, Gomez,
  Kaiser, and Polosukhin]{vaswani2017attention}
Ashish Vaswani, Noam Shazeer, Niki Parmar, Jakob Uszkoreit, Llion Jones,
  Aidan~N. Gomez, Lukasz Kaiser, and Illia Polosukhin.
\newblock Attention is all you need.
\newblock In \emph{NIPS}, pages 6000--6010, 2017.
\newblock URL \url{http://papers.nips.cc/paper/7181-attention-is-all-you-need}.

\bibitem[Good et~al.(1985)Good, Sweet, and Strobel]{good1985dynamic}
MC~Good, LM~Sweet, and KL~Strobel.
\newblock Dynamic models for control system design of integrated robot and
  drive systems.
\newblock 1985.

\bibitem[An and Hollerbach(1989)]{an1989role}
Chae~H An and John~M Hollerbach.
\newblock The role of dynamic models in cartesian force control of
  manipulators.
\newblock \emph{The International Journal of Robotics Research}, 8\penalty0
  (4):\penalty0 51--72, 1989.

\bibitem[Lynch and Park(2017)]{lynch2017modern}
K.M. Lynch and F.C. Park.
\newblock \emph{Modern Robotics: Mechanics, Planning, and Control}.
\newblock Cambridge University Press, 2017.
\newblock ISBN 9781316609842.
\newblock URL \url{https://books.google.de/books?id=8uS3AQAACAAJ}.

\bibitem[Hornik et~al.(1989)Hornik, Stinchcombe, and
  White]{hornik1989multilayer}
Kurt Hornik, Maxwell Stinchcombe, and Halbert White.
\newblock Multilayer feedforward networks are universal approximators.
\newblock \emph{Neural networks}, 2\penalty0 (5):\penalty0 359--366, 1989.

\bibitem[Wang et~al.(2016)Wang, Liu, Wu, Cao, Meng, and
  Kennedy]{wang2016training}
Shoujin Wang, Wei Liu, Jia Wu, Longbing Cao, Qinxue Meng, and Paul~J Kennedy.
\newblock Training deep neural networks on imbalanced data sets.
\newblock In \emph{2016 international joint conference on neural networks
  (IJCNN)}, pages 4368--4374. IEEE, 2016.

\bibitem[Geman et~al.(1992)Geman, Bienenstock, and Doursat]{geman1992neural}
Stuart Geman, Elie Bienenstock, and Ren{\'e} Doursat.
\newblock Neural networks and the bias/variance dilemma.
\newblock \emph{Neural computation}, 4\penalty0 (1):\penalty0 1--58, 1992.

\bibitem[Buda et~al.(2018)Buda, Maki, and Mazurowski]{buda2018systematic}
Mateusz Buda, Atsuto Maki, and Maciej~A Mazurowski.
\newblock A systematic study of the class imbalance problem in convolutional
  neural networks.
\newblock \emph{Neural Networks}, 106:\penalty0 249--259, 2018.

\bibitem[Koza and Koza(1992)]{koza1992genetic}
John~R Koza and John~R Koza.
\newblock \emph{Genetic programming: on the programming of computers by means
  of natural selection}, volume~1.
\newblock MIT press, 1992.

\bibitem[Bellman(1966)]{bellman1966dynamic}
Richard Bellman.
\newblock Dynamic programming.
\newblock \emph{Science}, 153\penalty0 (3731):\penalty0 34--37, 1966.

\bibitem[Rueckert et~al.(2017)Rueckert, Nakatenus, Tosatto, and
  Peters]{rueckert2017learning}
Elmar Rueckert, Moritz Nakatenus, Samuele Tosatto, and Jan Peters.
\newblock Learning inverse dynamics models in o (n) time with lstm networks.
\newblock In \emph{2017 IEEE-RAS 17th International Conference on Humanoid
  Robotics (Humanoids)}, pages 811--816. IEEE, 2017.

\bibitem[Cranmer et~al.(2020)Cranmer, Greydanus, Hoyer, Battaglia, Spergel, and
  Ho]{cranmer2020lagrangian}
Miles Cranmer, Sam Greydanus, Stephan Hoyer, Peter Battaglia, David Spergel,
  and Shirley Ho.
\newblock Lagrangian neural networks.
\newblock In \emph{ICLR 2020 Workshop on Integration of Deep Neural Models and
  Differential Equations}, 2020.
\newblock URL \url{https://openreview.net/forum?id=iE8tFa4Nq}.

\bibitem[Lutter et~al.(2019)Lutter, Ritter, and Peters]{lutter2019deep}
Michael Lutter, Christian Ritter, and Jan Peters.
\newblock Deep lagrangian networks: Using physics as model prior for deep
  learning.
\newblock In \emph{International Conference on Learning Representations}, 2019.
\newblock URL \url{https://openreview.net/forum?id=BklHpjCqKm}.

\bibitem[Greydanus et~al.(2019)Greydanus, Dzamba, and
  Yosinski]{NEURIPS2019_26cd8eca}
Samuel Greydanus, Misko Dzamba, and Jason Yosinski.
\newblock Hamiltonian neural networks.
\newblock In H.~Wallach, H.~Larochelle, A.~Beygelzimer, F.~d\textquotesingle
  Alch\'{e}-Buc, E.~Fox, and R.~Garnett, editors, \emph{Advances in Neural
  Information Processing Systems}, volume~32. Curran Associates, Inc., 2019.
\newblock URL
  \url{https://proceedings.neurips.cc/paper/2019/file/26cd8ecadce0d4efd6cc8a8725cbd1f8-Paper.pdf}.

\bibitem[Schmidt and Lipson(2009)]{Schmidt81}
Michael Schmidt and Hod Lipson.
\newblock Distilling free-form natural laws from experimental data.
\newblock \emph{Science}, 324\penalty0 (5923):\penalty0 81--85, 2009.
\newblock ISSN 0036-8075.
\newblock \doi{10.1126/science.1165893}.
\newblock URL \url{https://science.sciencemag.org/content/324/5923/81}.

\bibitem[Udrescu et~al.(2020)Udrescu, Tan, Feng, Neto, Wu, and
  Tegmark]{NEURIPS2020_33a854e2}
Silviu-Marian Udrescu, Andrew Tan, Jiahai Feng, Orisvaldo Neto, Tailin Wu, and
  Max Tegmark.
\newblock Ai feynman 2.0: Pareto-optimal symbolic regression exploiting graph
  modularity.
\newblock In H.~Larochelle, M.~Ranzato, R.~Hadsell, M.~F. Balcan, and H.~Lin,
  editors, \emph{Advances in Neural Information Processing Systems}, volume~33,
  pages 4860--4871. Curran Associates, Inc., 2020.
\newblock URL
  \url{https://proceedings.neurips.cc/paper/2020/file/33a854e247155d590883b93bca53848a-Paper.pdf}.

\bibitem[He et~al.(2016)He, Zhang, Ren, and Sun]{He_2016_CVPR}
Kaiming He, Xiangyu Zhang, Shaoqing Ren, and Jian Sun.
\newblock Deep residual learning for image recognition.
\newblock In \emph{Proceedings of the IEEE Conference on Computer Vision and
  Pattern Recognition (CVPR)}, June 2016.

\bibitem[Rifai et~al.(2011)Rifai, Vincent, Muller, Glorot, and
  Bengio]{rifai2011contractive}
Salah Rifai, Pascal Vincent, Xavier Muller, Xavier Glorot, and Yoshua Bengio.
\newblock Contractive auto-encoders: Explicit invariance during feature
  extraction.
\newblock In \emph{Icml}, 2011.

\bibitem[Ba et~al.(2016)Ba, Kiros, and Hinton]{ba2016layer}
Jimmy~Lei Ba, Jamie~Ryan Kiros, and Geoffrey~E Hinton.
\newblock Layer normalization.
\newblock \emph{arXiv preprint arXiv:1607.06450}, 2016.

\bibitem[Hochreiter and Schmidhuber(1997)]{HochSchm97}
Sepp Hochreiter and Jürgen Schmidhuber.
\newblock Long short-term memory.
\newblock \emph{Neural Computation}, 9\penalty0 (8):\penalty0 1735--1780, 1997.

\bibitem[Paszke et~al.(2019)Paszke, Gross, Massa, Lerer, Bradbury, Chanan,
  Killeen, Lin, Gimelshein, Antiga, Desmaison, Kopf, Yang, DeVito, Raison,
  Tejani, Chilamkurthy, Steiner, Fang, Bai, and Chintala]{NEURIPS2019_9015}
Adam Paszke, Sam Gross, Francisco Massa, Adam Lerer, James Bradbury, Gregory
  Chanan, Trevor Killeen, Zeming Lin, Natalia Gimelshein, Luca Antiga, Alban
  Desmaison, Andreas Kopf, Edward Yang, Zachary DeVito, Martin Raison, Alykhan
  Tejani, Sasank Chilamkurthy, Benoit Steiner, Lu~Fang, Junjie Bai, and Soumith
  Chintala.
\newblock Pytorch: An imperative style, high-performance deep learning library.
\newblock In H.~Wallach, H.~Larochelle, A.~Beygelzimer, F.~d\textquotesingle
  Alch\'{e}-Buc, E.~Fox, and R.~Garnett, editors, \emph{Advances in Neural
  Information Processing Systems 32}, pages 8024--8035. Curran Associates,
  Inc., 2019.
\newblock URL
  \url{http://papers.neurips.cc/paper/9015-pytorch-an-imperative-style-high-performance-deep-learning-library.pdf}.

\bibitem[Searson(2015)]{searson2015gptips}
Dominic~P Searson.
\newblock Gptips 2: an open-source software platform for symbolic data mining.
\newblock In \emph{Handbook of genetic programming applications}, pages
  551--573. Springer, 2015.

\bibitem[Gandomi and Alavi(2012)]{gandomi2012new}
Amir~Hossein Gandomi and Amir~Hossein Alavi.
\newblock A new multi-gene genetic programming approach to nonlinear system
  modeling. part i: materials and structural engineering problems.
\newblock \emph{Neural Computing and Applications}, 21\penalty0 (1):\penalty0
  171--187, 2012.

\bibitem[Kingma and Ba(2014)]{kingma2014adam}
Diederik~P Kingma and Jimmy Ba.
\newblock Adam: A method for stochastic optimization.
\newblock \emph{arXiv preprint arXiv:1412.6980}, 2014.

\end{thebibliography}
}

\end{document}